\pdfoutput=1

\documentclass[11pt]{article}

\usepackage[preprint]{acl}

\usepackage{times}
\usepackage{latexsym}

\usepackage[T1]{fontenc}

\usepackage[utf8]{inputenc}

\usepackage{microtype}

\usepackage{inconsolata}

\usepackage{graphicx}
\usepackage{multicol, multirow}
\usepackage{booktabs}

\usepackage{bbding}
\usepackage{pifont}
\usepackage{amssymb}

\usepackage{mathrsfs}
\usepackage{amsmath}
\usepackage{amsfonts}
\usepackage{bm}
\usepackage{bbm}
\usepackage{tabularx}
\usepackage{tablefootnote}
\usepackage{float}
\usepackage{xspace}
\usepackage{CJKutf8}
\usepackage{listings}
\usepackage{enumitem}
\usepackage[subrefformat=parens]{subcaption}
\usepackage{fontawesome5}
\usepackage{tcolorbox}
\usepackage{xcolor}
\newcommand{\update}[1]{#1}

\def\gecmetrics{\textsc{gec-metrics} }
\def\gecmetricsnospace{\textsc{gec-metrics}}
\def\gecmetricscomma{\textsc{gec-metrics}, }

\title{ \gecmetricsnospace: \\A Unified Library for Grammatical Error Correction Evaluation}

\author{
  \textbf{Takumi Goto}, \
  \textbf{Yusuke Sakai}, \
  \textbf{Taro Watanabe}
\\
  Nara Institute of Science and Technology (NAIST)
\\
  \texttt{\{goto.takumi.gv7, sakai.yusuke.sr9, taro\}@is.naist.jp}
}

\begin{document}
\maketitle
\begin{abstract}
We introduce \gecmetricsnospace, a library for using and developing grammatical error correction (GEC) evaluation metrics through a unified interface.
Our library enables fair system comparisons by ensuring that everyone conducts evaluations using a consistent implementation.
Moreover, it is designed with a strong focus on API usage, making it highly extensible. It also includes meta-evaluation functionalities and provides analysis and visualization scripts, contributing to developing GEC evaluation metrics.
Our code is released under the MIT license\footnote{\faGithub \ \ : \url{https://github.com/gotutiyan/gec-metrics}} and is also distributed as an installable package\footnote{\faPython \ \ : \href{https://pypi.org/project/gec-metrics/}{pip install gec-metrics}}. The video is available on YouTube\footnote{\faYoutube \ : \url{https://youtu.be/cor6dkN6EfI}}.
\end{abstract}

\section{Introduction}

Grammatical error correction (GEC) is a task that aims to automatically correct grammatical and surface-level errors, e.g., spelling, tense, expression, and so on~\cite{bryant-etal-2023-grammatical}. GEC serves as a writing support and is being successfully applied in commercial applications such as Grammarly. %
Therefore, many GEC methods have been proposed, such as sequence-to-sequence models~\cite{katsumata-komachi-2020-stronger, rothe-etal-2021-simple}, sequence labeling~\cite{awasthi-etal-2019-parallel, omelianchuk-etal-2020-gector}, and language model-based approaches~\cite{kaneko-okazaki-2023-reducing, loem-etal-2023-exploring}.
To evaluate their performance, some automatic GEC evaluation methods have been proposed (see Section~\ref{sec:related-work-evaluation-methods}). These evaluation methods are expected to exhibit a high correlation with human judgments, and their development has become an NLP task in itself.

\begin{figure}[t]
\centering
\includegraphics[width=\linewidth]{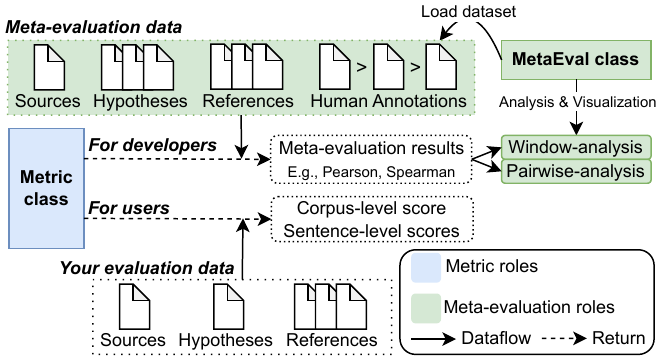}
  \caption{System overview of \gecmetricsnospace. The \emph{sources} are sentences containing grammatical errors, the \emph{hypotheses} are their corrected version, and the \emph{references} are human-corrected sentences. \texttt{Metric} classes support both corpus-level and sentence-level evaluation. The \texttt{MetaEval} classes conducts meta-evaluation of metrics, by calculating correlations with human evaluation. These classes also provide analysis and visualize scripts which are useful especially for developers.}
\label{fig:overview}
\end{figure}

Although various automatic GEC evaluation methods have been proposed, there is no common library that includes many of the latest studies, making it difficult to compare their performance. Indeed, this has caused several critical issues, such as unfair evaluation, high reproduction costs, and limited extensibility (see Section~\ref{subsec:problem}). In fact, most baseline scores are cited from reported results in previous studies, which makes it difficult to reproduce the original scores and to compare methods on new datasets or settings~\cite{maeda-etal-2022-impara}.

While GEC models are being unified through frameworks, UnifiedGEC~\cite{zhao-etal-2025-unifiedgec}, GEC evaluation metrics remain fragmented and lack a unified implementation, making consistent evaluation difficult. Model development and evaluation are inherently interconnected. For instance, the Hugging Face Transformers~\cite{wolf-etal-2020-transformers} has unified various language models into a single framework, while the Hugging Face Evaluate~\cite{von-werra-etal-2022-evaluate} has similarly consolidated evaluation metrics into a unified library, which has further accelerated and simplified model development. In the same way, a unified framework for the GEC evaluation metric is highly desired.

We introduce \gecmetricsnospace, a unified framework library that supports a variety of GEC evaluation metrics. It provides a unified interface with many useful features for comparison and developing new evaluation methods.
Figure~\ref{fig:overview} shows the workflow overview of \gecmetricsnospace.
In the figure, each module, i.e., ``Metric class'' and ``MetaEval class'', is easily extensible.
In addition, we carefully designed \gecmetrics to ensure transparency and reproducibility.
Furthermore, we provide a meta-evaluation interface that simplifies the development of new metrics. Our meta-evaluation experiments using the SEEDA~\cite{kobayashi-etal-2024-revisiting} dataset show that \gecmetrics can efficiently handle various evaluation metrics through a unified interface.

\section{Background}\label{sec:background}

\begin{figure}[t]
\centering
\includegraphics[width=\linewidth]{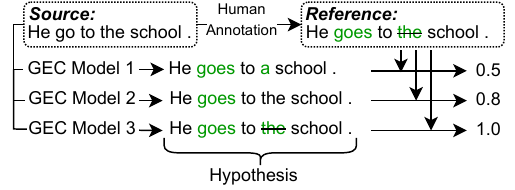}
  \caption{Examples of input/output for GEC evaluation.}
\label{fig:task}
\end{figure}

\subsection{Preliminaries for GEC Evaluation Metrics}\label{sec:background-metrics}

\begin{figure}[t]
\centering
\includegraphics[width=\linewidth]{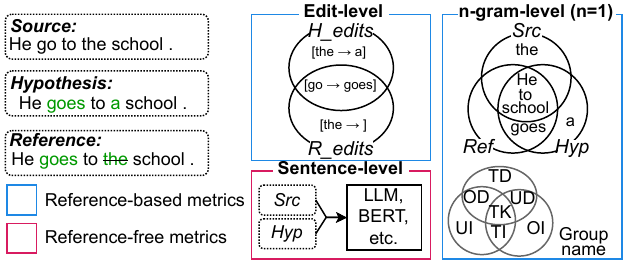}
  \caption{Categories of the current GEC metrics. The edit-level metrics considers the overlap of edits. The $n$-gram level metrics categorize $n$-gram into seven groups and use the $n$-gram count for each group. The sentence-level metrics employ neural models and estimate score without references.}
\label{fig:metric-category}
\end{figure}

Figure~\ref{fig:task} shows the overview of the GEC task and its evaluation. The source $S$ is a sentence containing grammatical errors, and hypothesis $H$ is its corrected version made by a GEC model: $H = \text{GECModel}(S) $. 
Basically, we also have one or more references $R$, which is a human-corrected sentence, for the evaluation.
The goal of the GEC evaluation is to assess the quality of the hypothesis.
The evaluation metrics are broadly categorized into reference-based and reference-free metrics, depending on whether they require references $R$.
\begin{equation}
    \text{Score} = \left\{
\begin{array}{ll} \text{Metric}(H | S, R)& \text{(Ref.-based)} \\
\text{Metric}(H | S) & \text{(Ref.-free)}
\end{array}
\right.
\end{equation}
\paragraph{Edit-level Metrics}
The reference-based metrics is often conducted by an edit-level evaluation.
The GEC field often handles sentence rewriting by decomposing into the granular level of editing.
By using automatic edit extraction method such as ERRANT~\cite{felice-etal-2016-automatic, bryant-etal-2017-automatic}, we extract two edit sets: hypothesis edit set $H_{edit}$ by comparing $S$ and $H$, and reference edit set $R_{edit}$ from $S$ and $R$. 
In Figure~\ref{fig:metric-category}, you can see there are two edits in each of $H_{edit}$ and $R_{edit}$. 
Then, we set the weight $w_e$ for each edit $e$, and calculate weighted scores: precision, recall, and $F_{\beta}$ score~\footnote{$F_{\beta} = \frac{\mathopen{}\left( 1 + \beta^{2} \mathclose{}\right) \text{Precision}\times \text{Recall}}{\beta^{2} \text{Precision} + \text{Recall}}$} by considering the intersection between $H_{edit}$ and $R_{edit}$: $I = (H_{edit} \cap R_{edit})$ in Equation (\ref{eq:fscore}).
For instance, a single edit [go → goes] is in both $H_{edit}$ and $R_{edit}$, thus $I = \{\text{[go → goes]\}}$ in Figure~\ref{fig:metric-category}.
\begin{equation}
\label{eq:fscore}
    \text{Precision} = \frac{
    \sum_{e \in I}w_e
    }{
    \sum_{e \in H_{edit}} w_e
    },
    \text{Recall} = \frac{
    \sum_{e \in I}w_e
    }{
    \sum_{e \in R_{edit}} w_e
    }
\end{equation}
\textbf{ERRANT}~\cite{felice-etal-2016-automatic, bryant-etal-2017-automatic} sets $w_e = 1.0$ for all of edits, and \textbf{PT-ERRANT}~\cite{gong-etal-2022-revisiting} computes a weight by BERTScore~\cite{zhang2019bertscore} or BARTScore~\cite{yuan2021bartscoreevaluatinggeneratedtext}.
\textbf{GoToScorer}~\cite{gotou-etal-2020-taking} uses the error correction difficulty, which is based on the correction success ratio of the predefined systems, as a weight\footnote{Precisely, the GoToScore additionally considers the non-corrected spans.}.

\paragraph{$n$-gram level Metrics}
The $n$-gram level metrics have also been employed for the reference-based evaluation.
~\citet{koyama-etal-2024-n-gram} provided a generic interpretation by an $n$-gram Venn diagram.
Figure~\ref{fig:metric-category} shows an example for $n=1$.
Each group in the Venn diagram is named as True Keep (TK), True Delete (TD), True Insert (TI), Over Delete (OD), Over Insert (OD), Under Delete (UD), Under Insert (UI).
In Figure~\ref{fig:metric-category}, you can see that \textit{He, to, school} are TK, \textit{the} is TD, \textit{a} is OI, and \textit{goes} is TI.
Similar to edit-based metrics, $n$-gram level metrics calculates precision or $F_{\beta}$ score from $n$-gram intersection.
\textbf{GLEU}~\cite{napoles-etal-2015-ground, napoles2016gleutuning} is a precision-based metric and \textbf{GREEN}~\cite{koyama-etal-2024-n-gram} uses $F_{\beta}$ score. Further detailed explanations are described in Appendix~\ref{sec:appen:detail-ngram}. 

\paragraph{Sentence-level Metrics}

Reference-free metrics are primarily designed as sentence-level metrics and are built using pretrained language models.
\textbf{SOME}~\cite{yoshimura-etal-2020-reference}  focuses on grammaticality, fluency, and meaning preservation; they fine-tuned BERT~\cite{devlin-etal-2019-bert} with regression head respectively optimize to human evaluation directly.
\textbf{Scribendi}~\cite{islam-magnani-2021-end} evaluates corrected sentences based on perplexity computed by a pretrained language model, and surface-level similarity.
\textbf{IMPARA}~\cite{maeda-etal-2022-impara} combines similarity scores between $S$ and $H$ with an quality estimation score for $H$. The quality estimation score is predicted using a BERT-based regression model trained to distinguish different levels of text quality.
\textbf{LLM-S}~\cite{kobayashi-etal-2024-large} performs 5-stage evaluation using a large language model. \textbf{LLM-E}~\cite{kobayashi-etal-2024-large} inputs edit sequences instead of corrected sentences.

\label{sec:related-work-evaluation-methods}

\subsection{Meta-Evaluation of GEC Metrics}\label{sec:background-meta}

The quality of GEC evaluation metrics is meta-evaluated by calculating the agreement between human evaluation results and metric-based evaluation results. 
Meta-evaluation is conducted from two perspectives: \emph{sentence-level} and \emph{system-level}.

In sentence-level evaluation, GEC evaluation methods score the hypothesis of multiple GEC systems associated with each source sentence.
Pairwise comparisons of hypotheses are performed for each source, and agreement between the human and metric evaluation results is accumulated over the entire data set. The reported scores are Accuracy (Acc.) and Kendall rank correlation coefficient ($\tau$).

In system-level evaluation, the focus is on comparing the overall relative quality of systems. System-level rankings are generally computed by averaging or accumulating sentence-level results. The metrics for system-level evaluation are Pearson ($r$) and Spearman ($\rho$) correlation coefficients.

To facilitate this, some meta-evaluation datasets have been proposed, such as GJG15~\cite{grundkiewicz-etal-2015-human} and SEEDA~\cite{kobayashi-etal-2024-revisiting}, which are derived from CoNLL-2014 shared task submissions~\cite{ng-etal-2014-conll}. Nonetheless, the number of available meta-evaluation datasets remains limited. One contributing factor is the lack of a unified framework for GEC evaluation metrics, which hinders consistent and comprehensive validation and increases the cost of implementing baselines when constructing meta-evaluation datasets.

\section{Problems of Existing Implementations}\label{subsec:problem}
\paragraph{Inconsistent interfaces.}

Although many GEC evaluation metrics have been proposed, their implementations are designed with their own interfaces and lack compatibility, such as input/output formats.
This makes cross-metric evaluation difficult and limits multifaceted discussions.
For example, recent evaluations of GEC model development heavily rely on ERRANT, while other metrics with high correlation to human evaluation, such as IMPARA, are seldom reported.
If the interfaces were unified, the complex experimental procedures caused by inconsistent implementations could be eliminated, which would facilitate better development and evaluation of GEC models.

\paragraph{Lack of official resources.}
\begin{table}[t]
    \centering
    \small
    \begin{tabular}{ll|cc}
    \toprule
    Metric & Reported paper & $r$ & $\rho$ \\
    \midrule
    Scribendi & \citet{islam-magnani-2021-end} & .951 & .940 \\
     & \citet{maeda-etal-2022-impara} & .303 & .729 \\
     & \citet{kobayashi-etal-2024-revisiting} & .890 & .923 \\
    \midrule
    IMPARA  & \citet{maeda-etal-2022-impara} & .974 & .934 \\
     & \citet{kobayashi-etal-2024-revisiting} & .961 & .965 \\
    \bottomrule
    \end{tabular}
    \caption{Previously reported meta-evaluation results on GJG15~\cite{grundkiewicz-etal-2015-human}. The r and $\rho$ are Pearson's correlation and Spearman rank correlation. The results  are inconsistent across studies, due to a lack of implementations and an open pre-trained model.}
    \label{tab:inconsistent}
\end{table}

Some metrics do not provide official resources. For example, Scribendi and LLM-\{S, E\} did not release their implementations, and IMPARA did not provide its fine-tuned weights. Therefore, we must reproduce these metrics, which can lead to discrepancies in reported results, as shown in Table~\ref{tab:inconsistent}.
Moreover, some metrics no longer work with their official code, such as GLEU, which is written in Python 2.
To avoid the cost of reproduction, most papers cite scores from previous studies, which compromises transparency.

\paragraph{API support.}

Since most original implementations are developed for specific experiments, they are typically intended to be executed using CLI-based scripts.  As a result, they do not support an extensible ecosystem such as APIs, which limits their flexibility and reusability.
When evaluation metrics are used as components in other methods, such as a reward function in reinforcement learning~\cite{sakaguchi-etal-2017-grammatical}, a utility function in MBR decoding~\cite{raina-gales-2023-minimum}, or a quality estimation model for ensembling~\cite{qorib-ng-2023-system}, APIs facilitate easier integration.

\section{\gecmetrics}

Our library, \gecmetricscomma compiles recent GEC evaluation methods into a unified interface. It supports not only the use of GEC metrics by users and GEC system developers but also meta-evaluation for GEC metric developers.
\gecmetrics supports both command-line usage and Python API access, enabling integration into a wide range of applications.
It resolves all the limitations of existing implementations highlighted in Section~\ref{subsec:problem}.
\update{ %
We have verified that the results obtained using \gecmetrics are consistent with those from official implementations for all publicly available metrics.
}

\subsection{Supported Methods}\label{subsec:supported}
\paragraph{GEC evaluation metrics.}
\gecmetrics supports all of ten metrics described in Section~\ref{sec:background-metrics}.
For reference-based metrics, it supports \textbf{ERRANT}, \textbf{PT-ERRANT}, and \textbf{GoToScorer} as edit-level metrics, \textbf{GLEU} and \textbf{GREEN} as $n$-gram level metrics. For reference-free metrics, it supports \textbf{SOME}, \textbf{Scribendi}, \textbf{IMPARA}, \textbf{LLM-S}, and \textbf{LLM-E}\footnote{Notably, our implementations of LLM-\{S, E\} are the first publicly available resource of \citet{kobayashi-etal-2024-large}. We contacted the authors, received some codes and prompts, and had several discussions to clarify the implementation details. We are deeply grateful for their support and contributions.} as sentence-level metrics. 
We carefully designed the library for extensibility and ease of changing hyper-parameters and base models, supporting various use cases such as modifying the value of $n$ in $n$-gram or switching the language models. Notably, LLM-\{S, E\} support the OpenAI and Gemini APIs, as well as all causal language models available in Hugging Face Transformers~\cite{wolf-etal-2020-transformers}, and also provides simplified prompts for applying to any data and scenario, as detailed in Appendix~\ref{sec:appen:llm-modifications}.

\paragraph{Meta-evaluation.}
\gecmetrics also supports all of two meta-evaluation frameworks: \textbf{GJG15} and \textbf{SEEDA} as introduced in Section~\ref{sec:background-meta}.  It accommodates all detailed configurations for each framework, ensuring comprehensive support.
Specifically, both datasets contain human Expected Wins~\cite{bojar-etal-2013-findings} rankings and human TrueSkill~\cite{NIPS2006_f44ee263} rankings. GJG15 adopts Expected Wins as the final human evaluation result, while SEEDA uses TrueSkill.
While system-level evaluation scores are typically reported using simple aggregation methods such as averaging, our library also provides the option to follow \citet{goto2025rethinkingevaluationmetricsgrammatical} by aggregating system-level results using either \textbf{\textit{Expected Wins}} or \textbf{\textit{TrueSkill}}.
Furthermore, SEEDA includes two evaluation settings: \textbf{SEEDA-S}, where human evaluation is conducted at the sentence level, and \textbf{SEEDA-E}, where evaluation is performed at the edit level. It also provides two configurations: \textbf{\texttt{Base}} and \textbf{\texttt{+Fluency}}. \gecmetrics fully supports all of these settings, enabling easy assessment of evaluation performance under diverse conditions.

\subsection{Interfaces}

\begin{lstlisting}[float=t, caption = An example of the implementation of evaluation and meta-evaluation\, using ERRANT as a metric and SEEDA as a meta-evalution framework., label = lst:api]
from gec_metrics.metrics import ERRANT
from gec_metrics.meta_eval import MetaEvalSEEDA
metric = ERRANT(ERRANT.Config(beta=0.5))
SRCS = ["He go to the school."] * 100
HYPS = ["He goes to the school."] * 100
REFS = [["He goes to school."] * 100]

# Corpus-level scoring
system_score: float = metric.score_corpus(
  sources=SRCS, hypotheses=HYPS, references=REFS
)  # Output: 0.833
# Sentence-level scoring
sent_score: list[float]  = metric.score_sentence(sources=SRCS, hypotheses=HYPS, references=REFS
)  # Output: [0.833, 0.833, ...]

### Meta-evaluation on SEEDA ###
meta = MetaEvalSEEDA(
  MetaEvalSEEDA.Config(system='base')
)
# System-level meta-evaluation
meta_system = meta.corr_system(metric)
print(f"SEEDA-S: {meta_system.ts_sent}")
# Output: MetaEvalBase.Corr(pearson=0.539, spearman=0.342)
# Sentence-level meta-evaluation
meta_sentence = meta.corr_sentence(metric)
print(f"SEEDA-S: {meta_sentence.sent}")
# Output:  MetaEvalBase.Corr(accuracy=0.594, kendall=0.188)
\end{lstlisting}

\begin{table*}[!t]
    \centering
    \small
    \setlength{\tabcolsep}{1.5pt}
    \resizebox{\linewidth}{!}{%
    \begin{tabular}{@{}l|rr|rrrr|rrrr|rr|rrrr|rrrr@{}}
    \toprule
     & \multicolumn{10}{c}{System-level} & \multicolumn{10}{c}{Sentence-level} \\
    \cmidrule(lr){2-11} \cmidrule(lr){12-21}
    Metric & \multicolumn{2}{c}{GJG15} & \multicolumn{4}{c}{SEEDA-S} & \multicolumn{4}{c}{SEEDA-E} & \multicolumn{2}{c}{GJG15} & \multicolumn{4}{c}{SEEDA-S} & \multicolumn{4}{c}{SEEDA-E} \\
     & \multicolumn{2}{c}{} & \multicolumn{2}{c}{Base} &  \multicolumn{2}{c}{+Fluency} &  \multicolumn{2}{c}{Base} &  \multicolumn{2}{c}{+Fluency} & \multicolumn{2}{c}{} & \multicolumn{2}{c}{Base} &  \multicolumn{2}{c}{+Fluency} &  \multicolumn{2}{c}{Base} &  \multicolumn{2}{c}{+Fluency}  \\
     & \multicolumn{1}{c}{$r$} & \multicolumn{1}{c}{$\rho$} & \multicolumn{1}{c}{$r$} & \multicolumn{1}{c}{$\rho$} & \multicolumn{1}{c}{$r$} & \multicolumn{1}{c}{$\rho$} & 
     \multicolumn{1}{c}{$r$} & \multicolumn{1}{c}{$\rho$} & \multicolumn{1}{c}{$r$} & \multicolumn{1}{c}{$\rho$} & \multicolumn{1}{c}{Acc.} & \multicolumn{1}{c}{$\tau$} & \multicolumn{1}{c}{Acc.} & \multicolumn{1}{c}{$\tau$} & 
     \multicolumn{1}{c}{Acc.} & \multicolumn{1}{c}{$\tau$} & \multicolumn{1}{c}{Acc.} & \multicolumn{1}{c}{$\tau$} & \multicolumn{1}{c}{Acc.} & \multicolumn{1}{c}{$\tau$} \\
     \midrule
     ERRANT &  .647 & .687 & .539 & .343 & -.592 & -.156 & .682 & .643 & -.508 & .033 & .654 & .307 & .594 & .189 & .544 & .087 & .608 & .217 & .558 & .116 \\
     PT-ERRANT &  .704 & .786 & .700 & .629 & -.548 & .077 & .788 & .874 & -.471 & .231 & .655 & .310 & .583 & .166 & .540 & .080 & .592 & .184 & .550 & .100 \\
     GoToScorer &  .668 & .615 & .726 & .601 & .439 & .499 & .816 & .762 & .514 & .635 & .579 & .159 & .550 & .100 & .511 & .021 & .563 & .126 & .524 & .048 \\
     GREEN &  .786 & .720 & \textbf{.925} & .881 & .185 & .569 & \underline{.932} & \underline{.965} & .252 & .618 & .660 & .319 & .600 & .199 & .552 & .105 & .574 & .148 & .537 & .073 \\
     GLEU &  .706 & .626 & .886 & \textbf{.902} & .155 & .543 & .912 & .944 & .232 & .569 & .673 & .346 & .672 & .343 & .616 & .231 & .673 & .347 & .625 & .249 \\
     SOME &  \textbf{.957} & \textbf{.923} & .892 & .867 & \textbf{.931} & .916 & .901 & .951 & \textbf{.943} & \underline{.969} & \textbf{.779} & \textbf{.559} & \textbf{.778} & \textbf{.555} & \textbf{.765} & \textbf{.531} & \textbf{.766} & \textbf{.532} & \textbf{.754} & \textbf{.509} \\
     IMPARA &  \underline{.956} & \underline{.885} & .916 & \textbf{.902} & \underline{.887} & \underline{.938} & .902 & \underline{.965} & \underline{.900} & \textbf{.978} & \underline{.747} & \underline{.495} & \underline{.753} & \underline{.506} & \underline{.738} & \underline{.475} & \underline{.752} & \underline{.504} & \underline{.743} & \underline{.486} \\
     Scribendi &  .855 & .835 & .620 & .636 & .604 & .714 & .825 & .839 & .715 & .842 & .728 & .457 & .660 & .320 & .623 & .245 & .672 & .345 & .648 & .295 \\
     \midrule
     GPT-4-E & .383 & .357 & .085 & .027 & -.817 & -.393 & .312 & .307 & -.764 & -.279 & .473 & -.053 & .520 & .041 & .582 & .165 & .538 & .077 & .591 & .183 \\
     GPT-4-S & -.073 & -.181 & .848 & .748 & .322 & .613 & .923 & .958 & .390 & .714 & .674 & .348 & .607 & .214 & .582 & .165 & .603 & .206 & .591 & .183 \\
     Gemini-S & -.205 & -.318 & .776 & .622 & .461 & .714 & .891 & .902 & .521 & .802 & .628 & .257 & .597 & .195 & .577 & .154 & .600 & .200 & .575 & .150 \\
     Qwen2.5-S & -.247 & -.274 & \underline{.920} & .839 & .788 & \textbf{.942} & .893 & .916 & .790 & .930 & .595 & .191 & .588 & .177 & .574 & .148 & .594 & .189 & .576 & .153 \\

     \midrule
     Ensemble  & .808 & .840 & .887 & .823 & .350 & .691 & \textbf{.953} & \textbf{.984} & .436 & .803 & -- & -- & -- & -- & -- & -- & -- & -- & -- & --\\
     (aboves w/o LLM) \\
    \bottomrule
    \end{tabular}
    }
    \caption{Meta-evaluation results using our \gecmetrics library. We use Pearson ($r$) and Spearman ($\rho$) for the system-level meta-evaluation, and accuracy (Acc.) and Kendall ($\tau$) for the sentence-level meta-evaluation. \textbf{Bold} is the highest value in each column, \underline{underline} is the second one.}
    \label{tab:results}
\end{table*}

\gecmetrics supports three types of interfaces: CLI, Python API, and GUI. While we primarily focus on the Python API, the other interfaces are demonstrated in Appendix~\ref{sec:appen:interfaces}.

Listing~\ref{lst:api} shows an example Python code for evaluation using ERRANT and meta-evaluation using SEEDA. 
Evaluation can be performed simply by passing a list of sentences to the \texttt{score\_**()} functions in L8 and L12. Similarly, meta-evaluation is supported through a simple interface, where the \texttt{corr\_**()} functions in L20 and L23 take a metric instance as input.
In addition, parameters and settings are separated via a \texttt{**.Config() dataclass}.
If switching to another metric, the process is simple and easy, thanks to the unified API interface.

\paragraph{Extensibility.} All classes are implemented by inheriting from an abstract class. The abstract class defines the minimal required methods, such as \texttt{score\_sentence()}, which must be overridden in the derived classes. This ensures that the interface remains consistent regardless of who implements the metric. Similarly, adding new meta-evaluation also requires only minimal implementation\footnote{We provide the documentation, including usage instructions, detailed API references, examples, and quick start guides: \url{https://gec-metrics.readthedocs.io/en/latest/index.html}.}.

\paragraph{Reproducibility.} CLI supports configuration input in YAML format. This allows users to share the exact settings used for running a metric, e.g., what model is used, contributing to high reproducibility.

\subsection{Analyses and Visualizations}
Meta-evaluation is not limited to correlation coefficients such as Pearson or Kendall but can also involve more detailed analyses.
\update{ %
For example, the \textit{window analysis}~\cite{kobayashi-etal-2024-revisiting} enables discussions on evaluation performance by focusing on competitive systems in human evaluation, and the \textit{edit-level attribution} shows which edit operation a metric focuses on in the evaluation~\cite{goto2024improvingexplainabilitysentencelevelmetrics}.
}
\gecmetrics provides tools for such analyses and result visualization.
\paragraph{Pairwise-analysis.}
Previous sentence-level meta-evaluations have primarily focused on Accuracy and Kendall’s $\tau$, which reflect overall agreement but offer limited interpretability. Therefore, we propose \emph{pairwise analysis}, which focuses on the relationship between differences in human rankings and agreement rates in sentence-level meta-evaluation. The difference between human- and metric-scored rankings for the same source can be calculated for each system pair, allowing agreement to be grouped and analyzed by ranking difference. Intuitively, the greater the difference in rankings assigned by humans, the more accurately a metric is expected to make judgments, reflecting how well it aligns with human evaluation at the sentence level.

\section{Experiments}

\paragraph{Settings.}\label{subsec:exp-settings}
Using our \gecmetrics library, we conducted meta-evaluations of GEC evaluation metrics. We employed all metrics listed in Section~\ref{subsec:supported}, and used GJG15 and SEEDA as meta-evaluation datasets. For system-level evaluation, we used the Expected Wins rankings from GJG15 and the TrueSkill rankings from SEEDA-\{S, E\}.
Appendix~\ref{sec:appen:metric-setup} provides the detailed experimental settings, which serve as the default configuration and generally follow those used in the original papers.
\begin{figure}[t]
\centering
\includegraphics[width=0.7\linewidth]{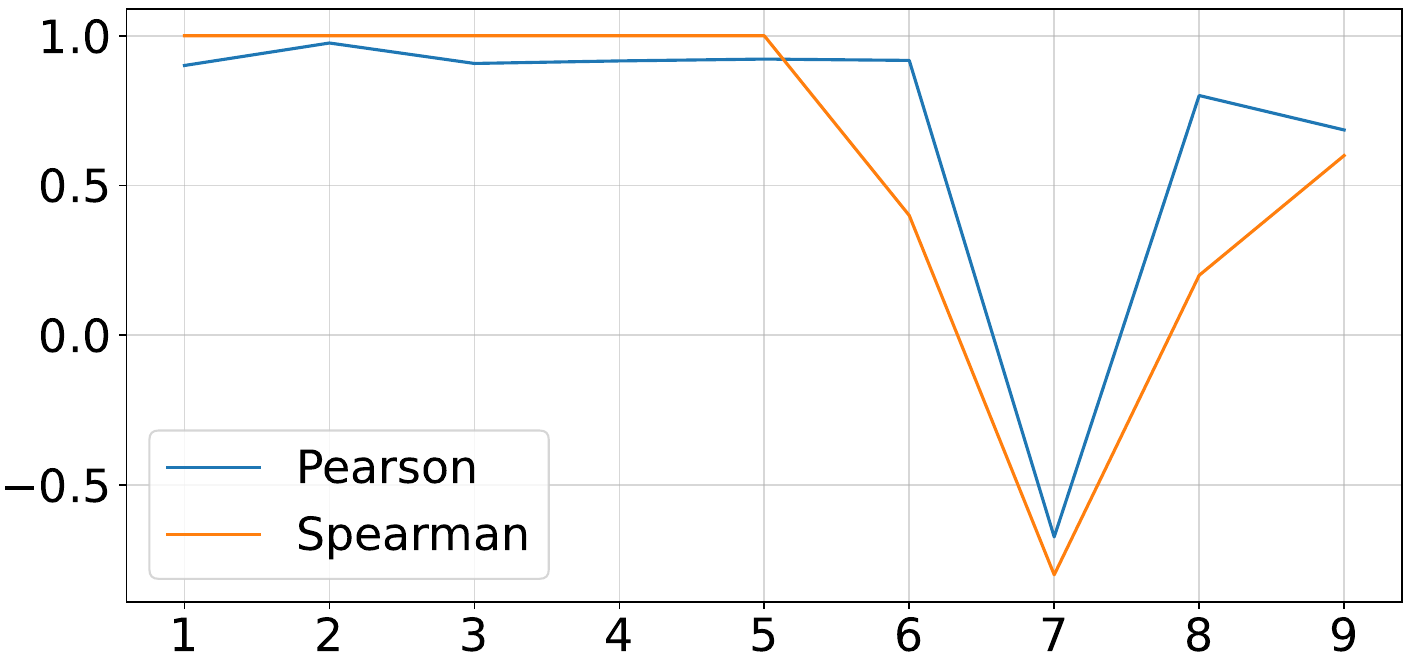}
  \caption{Window-analysis results for IMPARA. The x-axis indicates the start rank in the human-evaluation, and y-axis means Pearson (blue line) or Spearman (orange line) correlation.}
    \label{fig:window-impara}
\end{figure}
\paragraph{Extensive evaluation for LLM-\{S, E\}.}
We conducted several variations using different LLMs to provide extensive evaluation for LLM-\{S, E\}~\cite{kobayashi-etal-2024-large}.
Notably, we report results on GJG15 for the first time, revealing that the trend differs from SEEDA. 
Detailed motivations and settings are provided in Appendix~\ref{sec:appen:llm-modifications}.
We use \texttt{gpt-4o-mini-2024-07-18} (\textbf{GPT-4-S, GPT-4-E})~\cite{openai2024gpt4ocard}, \texttt{gemini-2.0-flash} (\textbf{Gemini-S})~\cite{geminiteam2025geminifamilyhighlycapable}, and \texttt{Qwen2.5-14B-Instruct} (\textbf{Qwen2.5-S})~\cite{qwen2.5} to emphasize extendability of our library for other language models.

\paragraph{Results.}

Table~\ref{tab:results} shows the experimental results.
ERRANT and PT-ERRANT show a higher correlation with SEEDA-E than with SEEDA-S, emphasizing the importance of aligning the evaluation granularity between human and automatic evaluations. Meanwhile, under the +Fluency setting, the correlation becomes negative, indicating the difficulty of evaluating GEC systems that focus on improving fluency. In contrast, SOME and IMPARA achieve high correlations even in the +Fluency setting. These results align with the trends reported in SEEDA~\cite{kobayashi-etal-2024-revisiting}.
On the other hand, for LLM-based metrics, while they achieve relatively high correlations in SEEDA, their performance is lower in GJG15. Our study is the first to apply LLM-based metrics to GJG15, suggesting that the evaluation capability of LLMs does not necessarily generalize and that there is room for improvement. Similarly, GPT-4-E fails to reproduce the results reported by ~\cite{kobayashi-etal-2024-large}, indicating the need for further discussion on the validity of the approach.
Figure~\ref{fig:window-impara} shows the window-analysis results for IMPARA.  We used human TrueSkill rankings of SEEDA-S and used 4 as the window size. An observation is that the correlations suddenly drops at $x=7$, which is consistent with \citeposs{kobayashi-etal-2024-revisiting} observation.

\paragraph{Metric Ensemble.}
GMEG-Metric~\cite{napoles-etal-2019-enabling} proposed an ensemble approach for evaluation metrics and reported robust performance across different domains. Given that new metrics continue to be developed after this work, ensemble techniques are expected to remain important for achieving reliable evaluations. Since ensembling requires results from multiple metrics, using a unified implementation like \gecmetrics facilitates experimentation. As a simple experiment to explore this, we consider using the average ranking across different metrics as the final evaluation score. For instance, if a system is ranked 2nd by a metric and 1st by another metric, its final evaluation score would be 1.5.
By ensembling metrics other than LLM-based metrics listed in Table~\ref{tab:results}, we achieved a Spearman rank correlation of 0.984 on SEEDA-E. This is the highest correlation in our experiment.
This short experiment shows that \gecmetrics facilitates the exploration of novel evaluation metrics.

\paragraph{Analysis for Sentence-level Scores.}\label{sec:pairwise-analysis}

\begin{figure}[t]
\centering
  \begin{minipage}[b]{0.23\textwidth}
    \centering
    \includegraphics[width=0.99\textwidth]{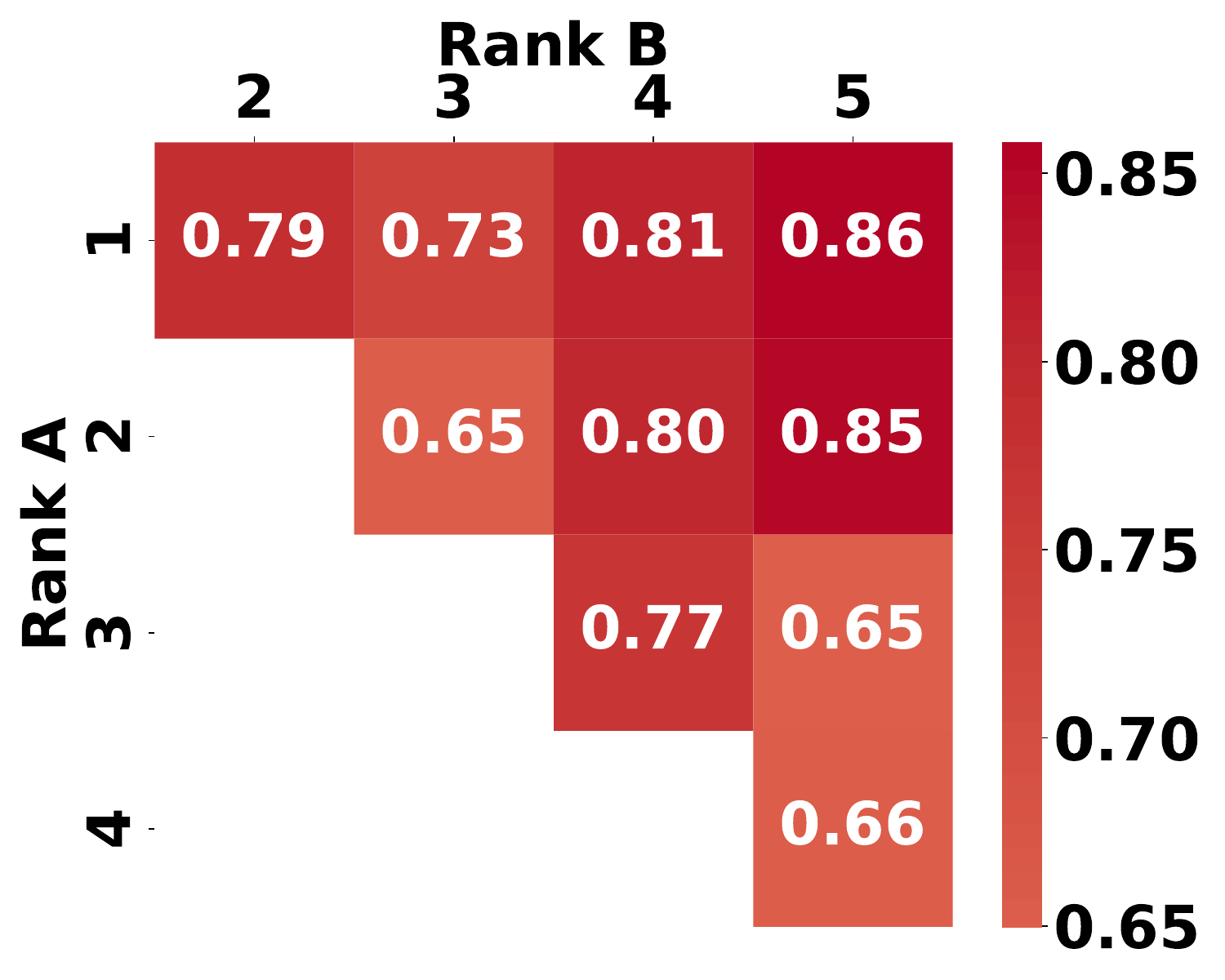}
    \subcaption{IMPARA}
    \label{fig:pair-impara}
  \end{minipage}
  \begin{minipage}[b]{0.23\textwidth}
    \centering
    \includegraphics[width=0.99\textwidth]{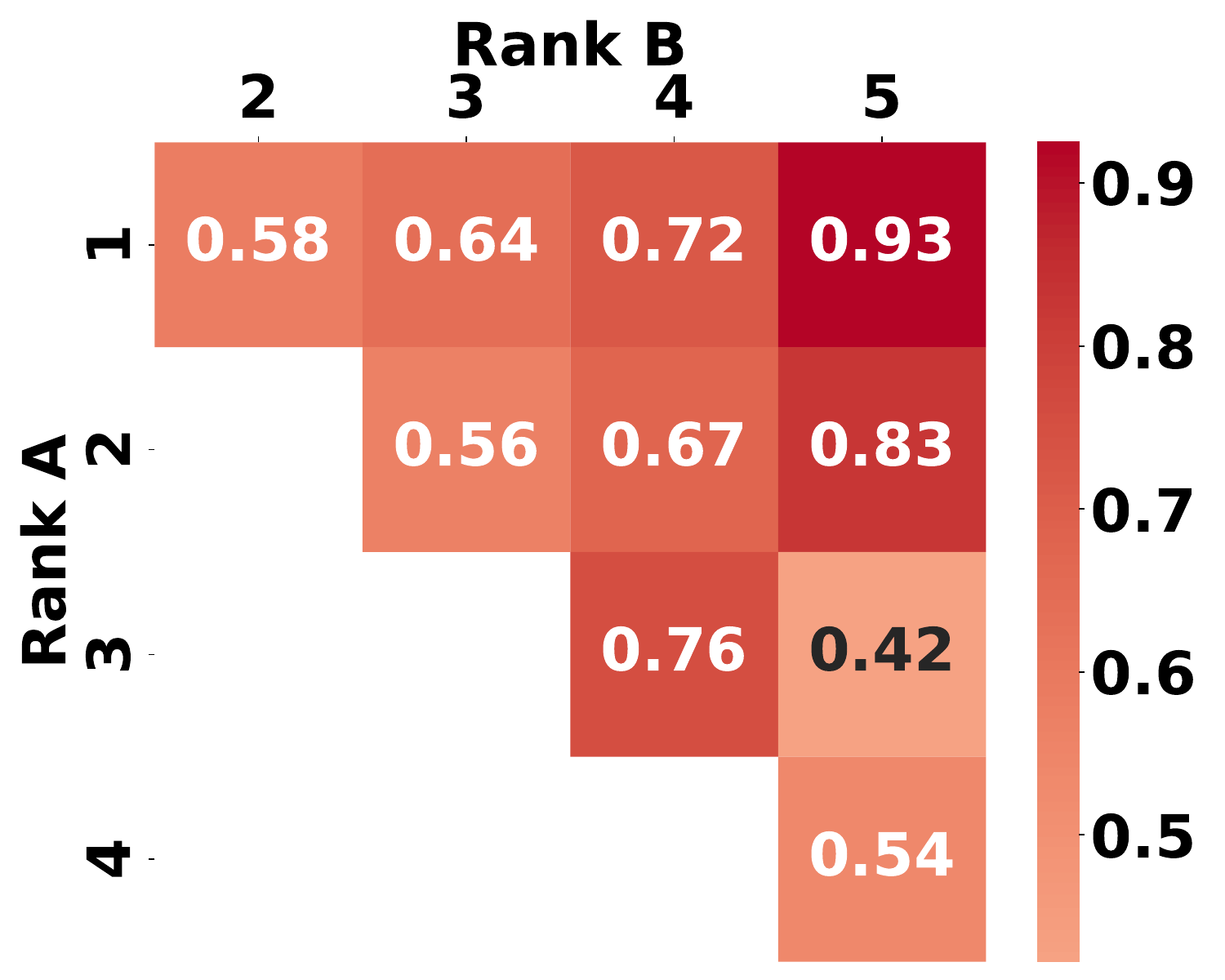}
    \subcaption{ERRANT}
    \label{fig:pair-errant}
  \end{minipage}
  \caption{Results of the pairwise-analysis. (a) shows the agreement rates between IMPARA and SEEDA-S annotation, and (b) shows the rates between ERRANT and SEEDA-E.}
    \label{fig:pair}
\end{figure}

Figure~\ref{fig:pair} presents the results of an experiment using human evaluation data from the SEEDA dataset. Rank A and Rank B correspond to the human-assigned rankings of a hypothesis pair.
Both of results are showing a trend where agreement increases as the difference in rankings grows (toward the upper right side in each figure).
This suggests that current metrics reflect human evaluative tendencies, but there is room for improvement in distinguishing minor differences in quality.

\section{Conclusion}
In this paper, we proposed a library, \gecmetricscomma to address issues in evaluation caused by inconsistencies in existing metric implementations and the lack of official resources. \gecmetrics is designed with a strong focus on API usability, making it easier to apply not only for evaluation but also for other purposes. Furthermore, it supports developers in improving evaluation metrics by providing an interface for meta-evaluation. 
We hope that our library will lead to further diverse applications and advanced research.
\update{ %
We will continue to develop our library, incorporating diverse methods and languages, and contribute to the community.
}

\section*{Ethics Statement and Broader Impact}

\paragraph{Contribution for research ethics.}
Using \gecmetrics improves the reproducibility and transparency of experiments, which is crucial from a research ethics standpoint. The inclusion of questions about implementation and experimental settings in the ACL Rolling Review checklist\footnote{\url{https://aclrollingreview.org/responsibleNLPresearch/}} highlights the community’s emphasis on these aspects. By continuing to maintain and develop metric implementations, \gecmetrics aims to support and strengthen these efforts.

\paragraph{Impacts for the community.}
\gecmetrics serves as a powerful tool for researchers to easily develop evaluation methods. It also accelerates their application in the GEC field, including bias investigations, integration with learning and inference methods such as reinforcement learning and ensembling, and use as a scorer in shared tasks. In fact, it has already been adopted as a scorer in a shared task competition at a domestic Japanese conference that examined metric vulnerabilities\footnote{\url{https://sites.google.com/view/nlp2025ws-langeval/task/gec}}. These cases demonstrate that \gecmetrics is beginning to contribute to advancing research.
At the same time, we recognize the importance of maintenance and management. We are committed to providing long-term support and actively incorporating new methods and pull requests responsibly.

\paragraph{License.}
We have also confirmed that there are no licensing issues with the code, methods, or data used in our implementation. 
\gecmetrics is released under the MIT license.

\section*{Acknowledgments}
\update{
We gratefully appreciate Masamune Kobayashi, the author of SEEDA and LLM-\{S, E\}~\cite{kobayashi-etal-2024-large, kobayashi-etal-2024-revisiting} for generously sharing code and prompts, as well as engaging in extended discussions, which served as a valuable reference during the development of our library.
We also thank the anonymous reviewers for their valuable comments and suggestions.
The architecture design of \gecmetrics is inspired by \textsc{mbrs}~\cite{deguchi-etal-2024-mbrs}.
This work has been supported by JST SPRING. Grant Number JPMJSP2140.
}

\bibliography{custom}

\appendix

\section{Details for \texorpdfstring{$n$}-gram level metrics.}\label{sec:appen:detail-ngram}

GLEU is a precision-based metric. By using the Venn diagram in the Figure~\ref{fig:metric-category}, it is formulated by:
\begin{equation}\label{eq:gleu}
    p_n = \frac{\text{TI}_n + \text{TK}_n - \text{UD}_n}{\text{TI}_n + \text{TK}_n + \text{OI}_n + \text{UD}_n}.
\end{equation}
Note that $\text{TI}_n, \text{TK}_n \dots$ represents the $n$-gram count of each group.
The $p_n$ is a precision for $n$-gram and is usually computed for each $n$ from 1 to 4. Then, the brevity penalty~\cite{papineni-etal-2002-bleu} is taken into account after taking the geometric mean.
\textbf{GREEN}~\cite{koyama-etal-2024-n-gram} is also an $n$-gram-level metric, but it computes the precision, recall, and $F_{\beta}$ score:
\begin{equation}
    \text{Precision}_n = \frac{\text{TI}_n + \text{TD}_n + \text{TK}_n}{\text{TI}_n + \text{TD}_n + \text{TK}_n + \text{OI}_n + \text{OD}_n},
\end{equation}
\begin{equation}
    \text{Recall}_n = \frac{\text{TI}_n + \text{TD}_n + \text{TK}_n}{\text{TI}_n + \text{TD}_n + \text{TK}_n + \text{UI}_n + \text{UD}_n},
\end{equation}
\begin{equation}
    F_{\beta} = \frac{\mathopen{}\left( 1 + \beta^{2} \mathclose{}\right) \text{Precision}\times \text{Recall}}{\beta^{2} \text{Precision} + \text{Recall}}.
\end{equation}
After calculating the geometric mean for each of precision and recall using $n$ from 1 to 4, the $F_{\beta}$ score is calculated.

\section{Details of experimental setup}\label{sec:appen:metric-setup}
For the reference-based metrics, we used the official two references of CoNLL-2014 shared task~\cite{ng-etal-2014-conll}. The below describes the detail exoerimental settings for each metric.
\begin{description}[leftmargin=0.2cm]
\item[ERRANT.] We use \texttt{errant==3.0.0}. Note that the extraction ways of edits have changed slightly between $\geq$v3.0.0 and $<$v3.0.0. We use $F_{0.5}$ as the score. The sentence-level scores are computed by choosing the best reference, which makes the highest $F_{0.5}$ score, for each source sentence.
\item[PT-ERRANT.] PT-ERRANT uses $F$-score of the BERTScore with \texttt{bert-base-uncased} for the edit-level weight computation. It rescales the weights by the baseline, but does not use the idf importance weighting. These are the same configurations as the official implementation\footnote{\url{https://github.com/pygongnlp/PT-M2}}. After computing edit-level weights, we compute weighed precision, recall, and $F_{0.5}$ score as in ERRANT. The computation method of the sentence-level scores is also the same as that of ERRANT.
\item[GoToScorer.] We used the first reference and all system outputs, including input sentences, for calculating the error correction difficulty.
\item[GLEU.] We use word-level GLEU and set 500 as the iteration count. The maximum $n$ is 4 for $n$-gram. The sentence-level scores are defined as the average of each reference.
\item[GREEN.] We use word-level GREEN and $F_{2.0}$.
\item[Scribendi.] We use GPT-2~\cite{radford2019language} as a language model to compute perplexity. The threshold for the maximum values of Levenshtein-distance ratio and token sort ratio is 0.8.
\item[SOME.] We use the official pre-trained weights, which are available from the official repository~\footnote{\url{https://github.com/kokeman/SOME}}. The weights for the grammaticality score, fluency score, and meaning preservation score are set to 0.55, 0.43, and 0.02, respectively.
\item[IMPARA.] For IMPARA, we reproduce the training experiments because no trained model is publicly available.
    As follows \citet{maeda-etal-2022-impara}, we generated 4,096 instances using CoNLL-2013~\cite{ng-etal-2013-conll} as the seed corpus, and split them into 8:1:1 for training, development, and evaluation sets.
    Thus, we used 3,276 instances as training data to fine-tune \texttt{bert-base-cased} and made public the pre-trained weights\footnote{\url{https://huggingface.co/gotutiyan/IMPARA-QE}}. \gecmetrics does not contain the training scripts, but we make them public in a separate repository\footnote{\url{https://github.com/gotutiyan/IMPARA}}. \texttt{bert-base-cased} is used for computing the similarity score with the threshold 0.9.
\item[LLM-S and LLM-E.] For GPT-4-S, we use \texttt{beta.chat.completions.parse} API for the OpenAI models and use \textsc{Outlines} library~\cite{willard2023efficientguidedgenerationlarge}\footnote{\url{https://github.com/dottxt-ai/outlines}} for the HuggingFace models, to ensure the output is in JSON structure. %
While \citet{kobayashi-etal-2024-large} uses \texttt{gpt-4-1106-preview}, we used \texttt{gpt-4o-mini-2024-07-18} model in our experiments to avoid using it due to the high experimental cost. We believe that not everyone can afford to use expensive models.
\end{description}

\section{Our Modifications of the LLM-based Metrics}\label{sec:appen:llm-modifications}
As described in Section~\ref{subsec:exp-settings}, we have made modifications to the LLM-based metric proposed by \citet{kobayashi-etal-2024-large}.
The first modification is the exclusion of contextual information from preceding and following sentences. Some datasets do not include surrounding context, and \citet{kobayashi-etal-2024-large} does not specify how to handle such cases. To ensure that evaluation is feasible for any dataset, we employed a prompt that does not incorporate contextual information, which also necessitated changes to the instruction text. We show the instruction text in Figure~\ref{fig:prompt:1}.

\begin{figure}[t]
\begin{tcolorbox}%
\small
The goal of this task is to rank the presented targets based on the quality of the sentences.

After reading the source sentence and target sentences, please assign a score from a minimum of 1 point to a maximum of 5 points to each target based on the quality of the sentence (note that you can assign the same score multiple times).

\# source

[SOURCE]

\# targets

... <omitted>
\end{tcolorbox}
\caption{Our modified instruction for LLM-S.}
\label{fig:prompt:1}
\end{figure}

The second modification clarifies the sampling method for input correction hypotheses. Their metric accepts up to five hypotheses simultaneously, but when evaluating a large number of systems, the number of different correction hypotheses may exceed five. In such cases, some method of selecting five sentences is required to proceed with evaluation. \citet{kobayashi-etal-2024-large} describes only the experimental setup for meta-evaluation using SEEDA, where pre-sampled correction hypotheses are used as input. However, this approach cannot be directly applied when evaluating a different set of systems or when working with a different dataset. Since \citet{kobayashi-etal-2024-large} does not define an experimental procedure for such scenarios, we adopted a method that selects five sentences based on their frequency, where frequency is defined as the number of systems that produce the same correction hypothesis. Note that multiple systems may output the same corrected sentence. The selected hypotheses are all unique, and the evaluation score assigned to each hypothesis is expanded across all systems that produced it. By selecting correction hypotheses with higher frequency, we maximize the number of systems that can be evaluated. We use a single RTX3090 for experiments.

\section{CLI and GUI Interfaces}\label{sec:appen:interfaces}

Listing~\ref{lst:cli-metric} provides an example of CLI. It can receive raw text files as inputs, the metric id to \texttt{-\,-metric}, and YAML-based configuration input using the \texttt{-\,-config} argument.

\begin{lstlisting}[language=sh, float=t, caption = Commandline usage of \gecmetrics. Each variable within < > indicates a path to a raw text file. You can use another metrics by specifying the - -metric argument\, e.g.\, ``- -metric impara''., label = lst:cli-metric]
gecmetrics-eval --src <src> \
    --hyps <hyp1> <hyp2> ... \
    --refs <ref1> <ref2> ... \
    --metric errant \
    --config config.yaml
\end{lstlisting}

\begin{figure}[!t]
\centering
  \begin{minipage}[b]{0.49\linewidth}
    \centering
    \includegraphics[width=0.99\linewidth]{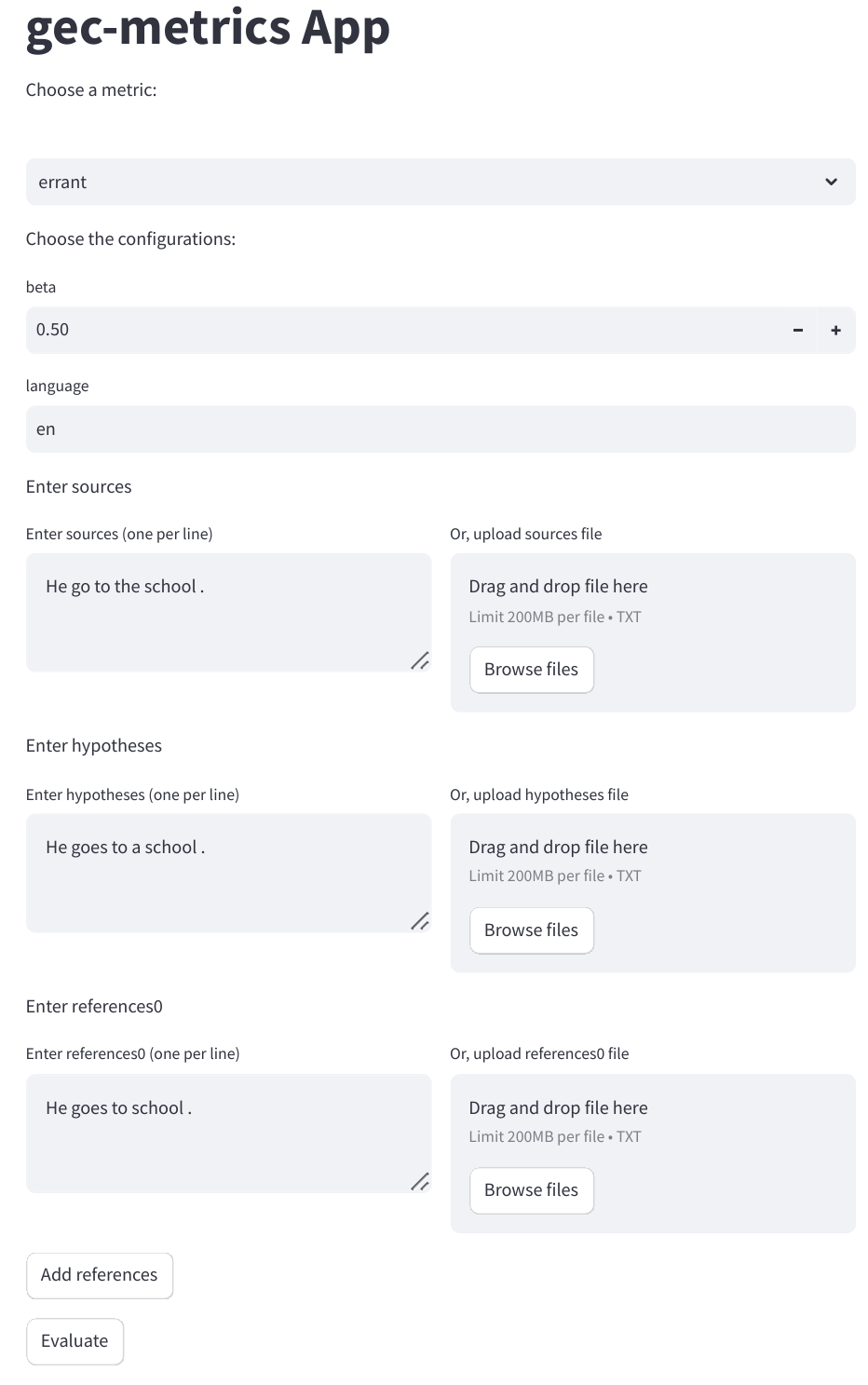}
    \subcaption{Metric GUI}
    \label{fig:app-metric}
  \end{minipage}
  \begin{minipage}[b]{0.49\linewidth}
    \centering
    \includegraphics[width=0.99\linewidth]{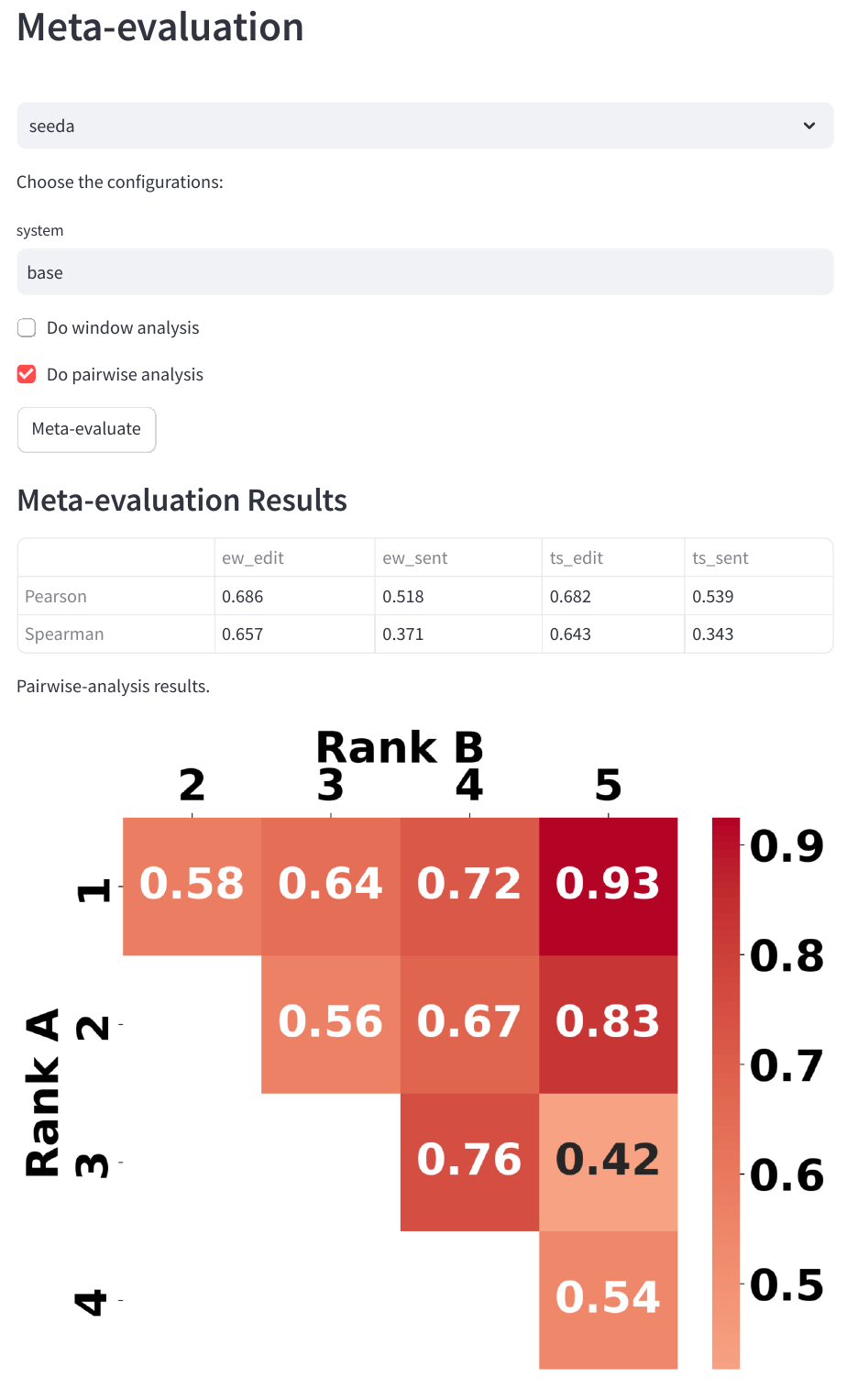}
    \subcaption{Meta-evaluation GUI}
    \label{fig:app-meta}
  \end{minipage}
  \caption{GUI of \gecmetrics. (a) is for metrics, and (b) is for meta-evaluation, which includes visualization of the analysis. They are actually combined on a single page.}
\label{fig:app}
\end{figure}

Figure~\ref{fig:app} shows a GUI example, which is developed via \textsc{Streamlit} library\footnote{\url{https://github.com/streamlit/streamlit}}. You can easily perform the evaluation for any dataset and the meta-evaluation, without coding. Furthermore, it has visualization features for the analysis results of meta-evaluation: window-analysis and pairwise-analysis, such as shown in Figure~\ref{fig:pair}. The code for GUI is provided in a separate repository: \url{https://github.com/gotutiyan/gec-metrics-app}.

\end{document}